\documentclass[twoside]{article}
\usepackage{iclr2016_conference,times}
\usepackage{hyperref}
\usepackage{url}
\usepackage{graphicx}
\usepackage{amsmath,amssymb,}
\usepackage{picinpar}
\usepackage[font=footnotesize]{caption}
\usepackage{subcaption}

\iclrfinalcopy

\newcommand{\x}{\boldsymbol{x}}
\newcommand{\z}{\boldsymbol{z}}
\renewcommand{\t}{\boldsymbol{t}}
\newcommand{\m}{\boldsymbol{m}}
\title{Bayesian Representation Learning\\ with Oracle Constraints}

\author{Theofanis~Karaletsos\\
Computational Biology, Sloan Kettering Institute\\
1275 York Avenue, New York, USA  \\
\texttt{Theofanis.Karaletsos@ratschlab.org} \\
\And
Serge Belongie\\
Cornell Tech\\
111 8$^{\mathrm{th}}$ Avenue \#302, New York, USA \\
\texttt{sjb344@cornell.edu} \\
\And
Gunnar R\"atsch\\ 
Computational Biology, Sloan Kettering Institute  \\
1275 York Avenue, New York, USA \\
\texttt{Gunnar.Ratsch@ratschlab.org} \\
}

\begin{document}

\maketitle

\vspace*{-1.2ex}
\begin{abstract}
Representation learning systems typically rely on massive amounts of labeled data in order to be trained to high accuracy. Recently, high-dimensional parametric models like neural networks have succeeded in building rich representations using either compressive, reconstructive or supervised criteria. However, the semantic structure inherent in observations is oftentimes lost in the process. Human perception excels at understanding semantics but cannot always be expressed in terms of labels. Thus, \emph{oracles} or \emph{human-in-the-loop systems}, for example crowdsourcing, are often employed to generate similarity constraints using an implicit similarity function encoded in human perception.
In this work we propose to combine \emph{generative unsupervised feature learning} with a \emph{probabilistic treatment of oracle information like triplets} in order to transfer implicit privileged oracle knowledge into explicit nonlinear Bayesian latent factor models of the observations.
We use a fast variational algorithm to learn the joint model and demonstrate applicability to a well-known image dataset.
We show how implicit triplet information can provide rich information to learn representations that outperform previous metric learning approaches as well as generative models without this side-information in a variety of predictive tasks.  In addition, we illustrate that the proposed approach compartmentalizes the latent spaces semantically which allows interpretation of the latent variables.
\end{abstract}

\vspace*{-1.5ex}
\section{Introduction}

Machine Learning excels in its ability to model large quantities of data with layered non-linear feature-learning systems for purposes such as classification and understanding of images, scenes, videos, text and more structured objects. 
Commonly, many successes are owed due to excessive availability of labels coupled with supervised learning. In other successful cases, the structure of the data is being used as a means to hard-code wiring for models, for instance modeling video using slowness and convolutions in images. 
Oftentimes, especially in the case of perception, a) the real structure of the data generating process is unknown and hard to explicitly model well, or b) large amounts of accurate labels are hard to come by or may even be inadequate for knowledge representation. 
One way to incoporate further information is to query oracles like crowds to gather cheap labels or to collect auxiliary information like similarity constraints accoring to undefined perceptual biases the crowd may be aware of. While labeling may be noisy or inadequate to represent knowledge, similarity constraints present a robust way to encode implicit information about various properties of stimuli.

We propose to take advantage of auxiliary (implicit) information provided by one or more \emph{oracles} as a means to learn flexible graphical models with latent variables. Examples of oracles include (human) crowds or implicit structural knowledge about the data, such as structural or multi-modal constraints without access to explicit features, which are encoded as triplet constraints (see Section~\ref{sec:triplet_model}).  Critically, we consider the oracle similarity constraints as implicit observations generated through an unknown process which we include in our model in order to capture subtle knowledge about similarity from the oracle(s). This key idea helps shape explicit, interpretable latent spaces that exceed the performance of purely unsupervised learning and can be applied in cases where labels are sparse or undesirable. These latent spaces can also be used to explicitly inspect the implicit knowledge passed on by the oracle to the model.
Our goal is to infer a latent factor model which learns jointly from triplets and observed data and transfers implicit biases encoded in the triplets into an explicit latent space that captures the semantics of the triplet-generating process better than simple density estimation (see Figure~\ref{fig_tripl:b}).  
We provide a detailed review of related work in Section~\ref{sec:relatedwork} where we explain the relationship of our model in the context of other triplet-loss based metric learning  approaches and generative models .

We first describe the two key contributions needed to perform the described task. In Section~\ref{sec:triplet_model} we introduce a novel probabilistic generative model of oracle observations. We extend this with a principled approach for multi-query oracles using masked subspaces in Section~\ref{sec:masked_oracle}. The second key contribution is described in Section~\ref{sec:opbn}, where we propose a principled approach combining the probabilistic oracle model with a graphical model performing nonlinear feature learning in order to transfer the implicit triplet knowledge into an explicit parametric model. In Section~\ref{sec:var_inf} we introduce a fast variational inference algorithm to learn posterior latent spaces respecting the observations of data and constraints.
Finally, in Sections~\ref{sec:results}~and~\ref{sec:discussion}, we present experimental results for benchmarking the proposed approaches, illustrating their properties and discussing the benefits they confer over competing approaches.


\vspace*{-1ex}
\section{Methods}
Let $\x \in \mathcal{R}^{N \times D}$ denote $N$ observations with $D$ dimensions.
We define latent variable $\z \in \mathcal{R}^{N \times H}$ corresponding to $H$-dimensional latent representations of datapoints $\x$.

\vspace*{-1ex}
\subsection{Probabilistic Modeling Of Oracle Triplets}
\label{sec:triplet_model}
We consider an unknown (dis)-similarity function $s_{Q}(\Phi(\x_i),\Phi(\x_j))$ that computes the distance between two objects $\x_i$ and $\x_j$ with respect to a query $Q$ based on semantic information associated with these two objects. 
We consider $\z=\Phi(\x)$ to be the internal conceptual representation the oracle uses to apply similarity function $s_{Q}(\cdot,\cdot)$ for an unobserved feature space $\Phi$. In addition, we consider the case where we can not directly observe the similarity function, but were we only observe orderings over similarities of $\z_i$ to $\z_j$ and $\z_l$, i.e., either $s_Q(\z_i, \z_j)$ is greater (equal) or smaller than $s_Q(\z_i, \z_l)$. 
We define the set of all oracle triplets related to query $Q$ as:
\begin{align}
\label{triplet_definition}
\mathcal{T_{Q}}=\{(i,j,l)~~|~~s_{Q}(\z_i,\z_j) > s_{Q}(\z_i,\z_l) \}.
\end{align}
We do not have access to the exhaustive set $\mathcal{T_{Q}}$, but can sample $K$-times from it using the oracle to yield a finite sample $\mathcal{T_{Q}}^{K}=\{t_k\}_{k=1}^{K}$.

An illustrative example of this process is human perceptual judgement of similarities, which heavily relies on internal representations and abstracted concepts to evaluate similarities over purely using raw low level image statistics.
A frequently used oracle is the crowd. Systems like Amazon Mechanical Turk are used to obtain triplet samples to explore the human perceptual prior as an oracle. However, oracles also naturally arize from data structure, such as temporal or spatial orderings or other known semantic structure. Another type of oracle is access to privileged information, which are extra features only implicitly available through the triplets, such as sentiments associated with visual features. A shared property of all these oracles is that they provide weak natural constraints on similarity without explicitly quantifying it.

We model the likelihood $t_{i,j,l}$ of a triplet being contained $\mathcal{T_{Q}}$ as a draw from a Bernoulli distribution over the states \emph{True} and \emph{False} parametrized using a softmax-function. If we consider 
\begin{equation}
p(t_{i,j,l})=\int\limits_z p(t_{i,j,l}|z_i,z_j,z_l) p(\z_{i}) p(\z_{j})p(\z_{k}) dz_{i}dz_{j}dz_{k},
\end{equation}this gives the following likelihood:\vspace*{-0.6ex}
\begin{equation}
\begin{split}
p(t_{i,j,l}) = Ber(t_{i,j,l}) = \frac{\text{e}^{-D_{i,j}}}{\text{e}^{-D_{i,j}}+\text{e}^{-D_{i,l}}}
\end{split}
\label{eq_lk_crowd}
\end{equation}
with \vspace*{-1ex}
\begin{equation}
D_{a,b} = \sum \limits_{h=1}^{H} D^{h}_{a,b}=-\sum \limits_{h=1}^{H} \left[ \text{JS}\Big(p(\z_{a}^{h})||p(\z_{b}^{h})\Big)\right].
\label{eq_D12}
\end{equation}

and $\text{JS}\Big(p(\z_{a}^{h})||p(\z_{b}^{h})\Big)=\frac{1}{2} 
\text{KL}\Big(p(\z_{a}^{h})||p_{q}(\z^{h})\Big) +\frac{1}{2} \text{KL}\Big(p(\z_{b}^{h})||p_{q}(\z^{h})\Big)$ with $p_{q}(\z^{h}) = \frac{1}{2} p(\z_{a}^{h}) + \frac{1}{2} p(\z_{b}^{h})$ and $\text{KL}(q || p)=\int \limits_{\z} q(\z) \text{log}(\frac{q(\z)}{p(\z)}) d\z$.

We denote the likelihood in \eqref{eq_lk_crowd} as {\bf BER} in the rest of the paper. Since the Jensen Shannon divergence above is commonly intractable, we discuss alternatives used in our experiments in Supplement~\ref{sec:appB}.

A subtlety of the acquisition process for the triplets is that oracles are not asked to provide distances, but just binary statements over similarity rankings based on the prompted question. Thus, any triplet which fulfills the statement made by the definition is valid. We model this relaxation into a \emph{truncated} Bernoulli likelihood as $p(\t_{i,j,l}) = \begin{cases} 1 &\mbox{if } Ber(\t_{i,j,l}) \geq 0.5 \\ 
Ber(\t_{i,j,l}) & \mbox{if } Ber(\t_{i,j,l}) < 0.5 \end{cases}$ and refer to it as {\bf TBER} later.\vspace*{-1ex}
%
%

\subsection{Masked Oracle Models}
\label{sec:masked_oracle}

\begin{figwindow}[4,r,{\includegraphics[width = 2.9in]{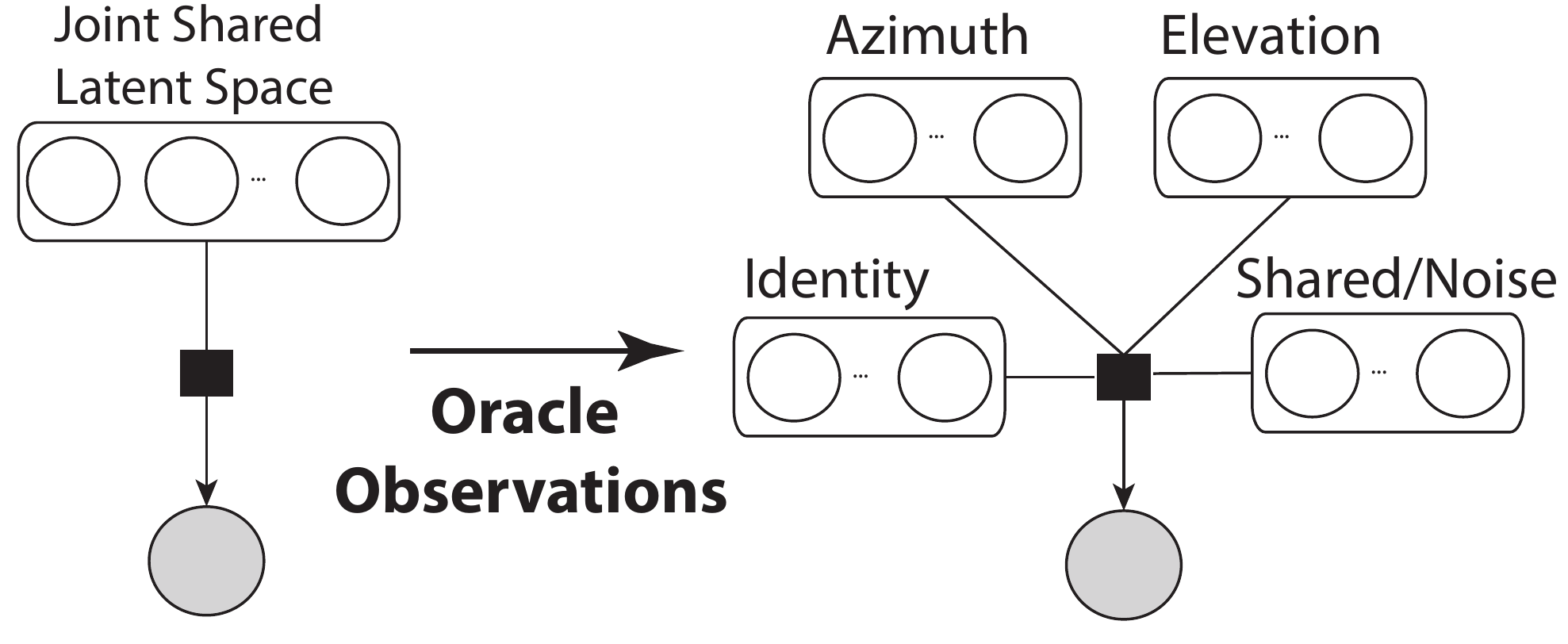}
},{Conceptually, observing oracle information from multiple oracles/questions allows an otherwise fully unsupervised and unstructured model (left) to identify a semantically compartmentalized generative process by using masked subspaces (right). In Results we model images of illuminated faces and show that by using oracle observations the latent space factorizes automatically into subspaces related to different semantic aspects.\label{fig_systemid}}]
Frequently, oracle information can be conflicting, especially when using multiple oracles. For instance, Consider colored geometric shapes, where we have a red triangle, a red circle and a blue circle. If we ask one oracle to compare shapes with a set of triplets and we furthermore ask another oracle to compare colors, we may get conflicting oracle constraints. The circles are more similar shapes while the red circle and triangle have more similar colors. The generated oracle constraints cannot easily be jointly fulfilled with uniform global constraints. We extend the presented model of oracle observations by incorporating \emph{masks over the dimensions of the latent space}, which weigh/select dimensions on which the oracle constraints must hold. Since latent variables typically are of higher dimension than 1, this approach aims at learning semantic latent subspaces where specific variables encode features which correspond to semantic information associated with an oracle. These subspaces can in general be entirely private to an oracle question, or may share information with multiple questions. This leads to a compartmentalization of the semantic representation. In Figure~\ref{fig_systemid} we give another conceptual illustration and example that we also use in Results.
\end{figwindow}
Formally, for a set of $H$-dimensional latent variables $\z$ we define a corresponding H-dimensional global mask-variable $\m^Q$ which is shared between all samples and is specific to a question/oracle $Q$. Using these masks, we adapt \eqref{eq_D12} to yield the {\it masked oracle model}:
\begin{equation}
D^{\m}_{di,j}= \sum \limits_{h=1}^{H} m_{h} D^{h}_{i,j}.
\label{eq_dij}
\end{equation}
%
%
We define learned masks by $m_h=\sigma(b_h)$, where $b_{h}\sim \mathcal{N}(0,1)$ and $\sigma$ denotes the sigmoid function.

\subsection{Variational Belief Networks}
\label{section:vae}
Apart from modeling oracle triplets defined over latent representations $\z\in\mathcal{R}^H$, we are interested in modeling observations $\x\in\mathcal{R}^D$ well.
We learn a graphical model to maximize $p(\x)$ using $H$-dimensional latent variables $\z$. The latent variables can be drawn from any exponential family distribution $p(\z)$, but simplifying cases for inference and learning exist for many continuous distributions. We can write the model as follows:
\begin{align}
p(\x ;\theta) = \int\limits_{\z} p_{\theta}(\x|\z) p(\z) d{\z},
\end{align}
with $p_{\theta}(\x|\z) := f(\x; \z,\theta)$ being an exponential family likelihood with parameters given as a function of $\z$. Also, $f$ is a function parametrized by $\theta$ (for instance, a multi-layer perceptron). 
Here, we focus on the Gaussian distribution as a prior $p(\z)$ with dimensions $H$, but note that with small adaptations to the inference procedure other distributions are feasible.
In this case we predict $D$ means $\mu_{\z}$ and variances $\sigma^{2}_{\z}$ for each dimension $d$ given the state of latent variables $\z$ using $\mu_{\z,d} = f_{\theta,\mu,d}(\z)$ and $\log\,\sigma_{\z,d} = f_{\theta,\sigma,d}(\z)$. 

 A good estimator for learning the parameters of such a model by assuming an approximate conditional posterior $q_{\phi}(\z|\x)$ was suggested in~\cite{kingma2013auto,rezende2014stochastic,mnih2014neural}, all of which can be understood as instances of doubly stochastic variational inference.
The estimator forms a variational lower bound~\cite{wainwright2008graphical,jordan1999introduction} to the marginal likelihood. Performing coordinate ascent with respect to variational parameters $\theta$ \& $\phi$ corresponds to minimizing the divergence between the true and the approximate posterior:
\begin{align}
\text{log}\, p_{\theta}(\x^{(i)}) \geq \mathcal{L}(\theta,\phi;\x^{(i)})=-\text{KL}(q_{\phi}(\z)|| p_{\theta}(\z)) + \text{E}_{q_{\phi}(\z)} [ \text{log}\, p_{\theta}(\x^{(i)}| \z )].
\end{align}
In the following, we will refer to this fully unsupervised model as {\bf VAE}.

\subsection{Joint model of triplets and observations: Oracle-Prioritized Belief Network}
\label{sec:opbn}

After having established an observation model for triplets $\t$ and an observation model for $\x$, we can proceed to introduce the full generative process for a joint model over both observables.
Instead of relying on a supervised model taking observations $\x$ as input, we prefer using a generative approach that models the joint density of both data and triplets to provide an unsupervised model which requires only input data and samples from an oracle to be trained. \vspace*{-2ex}

\begin{figwindow}[0,r,{\includegraphics[width = 2.9in]{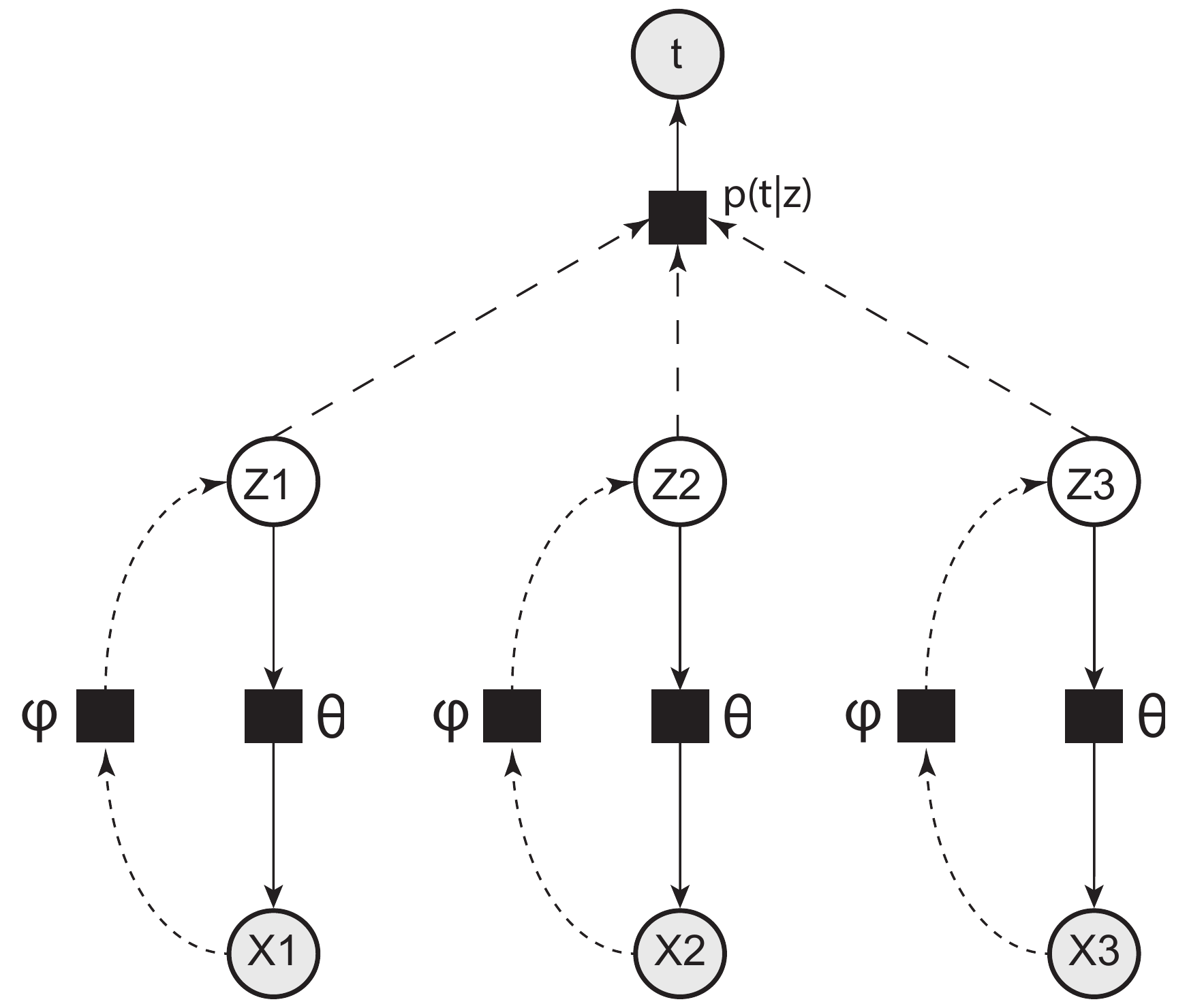}
},{Shown is the proposed joint model. It models observations $\x$ and triplets from an oracle $\t$ as observed (shaded) variables. The latent space with variables $\z$ \emph{causes} the shaded variables and thus captures the information necessary for modeling both.\label{fig_tripl:b}}]

The advantage is that the generative latent variable encoding does not throw away information about the observations, leading to models which explicitly need to capture latent factors generating the data. When observing oracle-samples, learning leads to explicit factors of the information from both the oracle and the observations.

We use a belief network as introduced in Section~\ref{section:vae} to model observations $\x$ and connect the latent variables with an oracle observation-term for triplets $\t$ introduced in Section~\ref{sec:triplet_model}. Triplets require multiple samples from the prior to be drawn, as they are defined over multiple objects jointly. Similar to the inference model in the Siamese network~\citep{chopra2005learning}, this necessitates multiple instances of the model with shared parameters to work in coordination to generate a triplet. We sketch the generative model in Figure~\ref{fig_tripl:b}.
%
\end{figwindow}

For our proposed joint-model that we will refer to as {\bf OPBN}, we consider $N$ datapoints and $K$ triplets defined over them:\vspace*{-1ex}
\begin{align}
\label{crowd_BN}
p(\x,{\bf t};\theta)= \int \limits_{\z} \prod \limits_{n}^{N} \left[ p( \z_n) p_{\theta}( \x_n|\z_n)  \right] \prod \limits_{k}^{K} \left[ p(t_k| \z_{k_i},\z_{k_j},\z_{k_l} ) \right] dz
\end{align}
The generative process according to this model is:
\begin{enumerate}
\item sample $\z_i$, $\z_j$, $\z_l$ from prior: $\z\sim \mathcal{N}({\bf 0},{\bf 1})$;
\item for each $\z$, sample observation $\x$ using nonlinear likelihood, e.g. $p(\x|\z)=\mathcal{N}(\x;{\bf \mu}_{\z}, {\bf \sigma}^{2}_{\z})$;
\item for each set $\{i,j,l\}$, sample triplet $\t_{i,j,l}\sim p(\t|\z_i,\z_j,\z_l)$.
\end{enumerate}

Triplets tie together multiple datapoints and capture their dependencies through the latent representations. This has the effect of attaching higher-order potentials to the latent space, which the model uses for regularization and guidance.
It is noteworthy that learning consists of maximizing the marginal likelihood $p(\x,\t)$ by integrating out the latent $\z$'s.
This directly maximizes the evidence coming from the oracle and the observations, while maintaining flexibility for the model used in-between. This model balances a reconstruction cost for the datapoints, the generative cost for the triplets and the prior on the latent variables when generating samples.


\subsection{Learning using fast Variational Inference}
\label{sec:var_inf}
Our goal is to maximize the marginal likelihood of the evidence, $\text{log}p_{\theta}(\x,\t)$ in order to learn a good mapping capturing the dependencies between observations $\x$ and triplets $\t$. This involves integrating out the latent variables which is in general analytically intractable in highly flexible model classes.
In order to perform efficient learning and inference in the model given by \eqref{crowd_BN} we resort to approximate inference methods and employ doubly stochastic Variational Inference~\citep{kingma2013auto,rezende2014stochastic,titsias2014doubly}. Variational Inference~\citep{jordan1999introduction,wainwright2008graphical} requires approximate distributions $q(\z)$ over the posterior of the latent variables. We use amortized inference by employing an inference network to learn a conditional variational distribution $q_{\phi}(\z|\x)$ with an MLP parametrized by $\phi$. The inference model predicts the variational approximation to the posterior latent variables per input data point. The evidence lower bound (ELBO) looks as follows:
\begin{equation*}
\begin{split}
\label{var_inf_crowd}
\text{log}\,p_{\theta}(\x,\t) =& \mathcal{L}(\theta,\phi;\x,\t) + \text{KL}(q(\z)||p(\z|\x,\t)) \geq  \mathcal{L}(\theta,\phi;\x,\t)\\
= & - \mathbb{E}_{n} \Big[ \text{KL}(q(\z^n)|| p_{\theta}(\z)) \Big]
   + \mathbb{E}_{n} \Big[ \mathbb{E}_{q(\z)} [ \text{log}\,p_{\theta}(\x^n| \z^n )]\Big] 
   + \mathbb{E}_{k} \Big[ \mathbb{E}_{q(\z)} [ \text{log}\,p(\t^k| \z^{k_{ijl}})]\Big]\vspace*{-0.5ex}
\end{split}
\end{equation*}
where $k_{ijl}$ acts as an index on matrix $\z$ selecting the corresponding datapoints.
Theoretically, performing coordinate ascend on this lower bound is sufficient to infer the parameters of the model $\theta$ and inference network $\phi$.
However, the expectations over latent variables $q(\z)$ present in the ELBO are intractable. We resort to the reparametrization trick~\citep{kingma2013auto,rezende2014stochastic,titsias2014doubly} and perform doubly stochastic variational inference by drawing $L$ unbiased samples $\z^l$ from these expectations using the identity $\z^l = \mu_{\phi} + \lambda_{\phi} \cdot \epsilon^l$, where $\{\mu_{\phi},\lambda_{\phi}\}$ are predicted variational parameters using the inference network and $\epsilon^l \sim \mathcal{N}(0,1)$ are unbiased samples from a unit Gaussian. The differentiable new bound $\mathcal{L}(\theta,\phi;\x,\t)$ then takes the shape: 
%
$ -\mathbb{E}_{n} \Big[ \text{KL}(q_{\phi}(\z^n|\x^{n})|| p_{\theta}(\z)) \Big]
 + \mathbb{E}_{n} \Big[ \frac{1}{L}\sum\limits_{l=1}^{L} [ \text{log} p_{\theta}(\x^n| {\bf z^l} )]\Big]
 + \mathbb{E}_{k} \Big[ \frac{1}{L}\sum\limits_{l=1}^{L} [ \text{log} p(t^k| {\bf z^{k_{ijl}^{l}}})]\Big]
$. 
%
On this new objective we can now perform gradient-based learning by following $\nabla{\mathcal{L}(\theta,\phi;\x,{\bf t})}$ with respect to global variational parameters $\phi$, $\theta$. We perform stochastic gradient descent by drawing minibatches with $N_b$ datapoints and $K_b$ triplets each time.

When learning masks ${\bf m}$ as in Section~\ref{sec:masked_oracle}, we infer posterior distribution of the masks given observed data $p({\bf m}|\mathcal{D})$. We use variational inference analogously to learn an approximate distribution $q({\bf m}^{h};\xi)=\mathcal{N}(\mu_{h}^{\xi},\sigma_{h}^{\xi})$ by adding the KL loss for the masks to the ELBO while taking into account the state of the mask variable in the triplet likelihood. 

Upon close inspection we detect that the components $q_{\phi}(\z|\x)$ and $p_{\theta}(\x| \z)$ form a variational autoencoder where the parameters have distilled the triplet information. This also clarifies where the transfer of implicit information from the triplets to the learned parametric model happens. In simple terms, the formulation of the model forces the inference network to learn encodings respecting the triplets and the model $p_{\theta}( \x|\z)$ decodings which account for that shared information.

\vspace*{-1ex}
\section{Relationship to other work}\vspace*{-1ex}
\label{sec:relatedwork}

Similarity-based learning, for instance via crowdsourcing, has been tackled in various ways in the community before.
Notably, crowd-kernels are inferred and used for various vision tasks using~\cite{van2012stochastic} which assumes a fixed Student-t structure to produce an embedding using similarity constraints from a crowd, but does not learn an adaptive latent representation of the input features.
In~\cite{chechik2010large}, a metric respecting the particular distances in similarity is learned. This differs from the case we are studying, as it assumes that specific distances or similarities are observed, which is hard to ask of a weak oracle.
In~\cite{tamuz2011adaptively}, a probabilistic treatment for triplets is introduced and an adaptive crowd kernel is learned without specific visual features in mind. While we also adopt a probabilistic treatment of triplets, we will learn an adaptive feature representation comparing images from the crowd as well.

Flexible nonlinear models have been employed in a variety of situations to learn representations for data.
A key result in relation to this work is the Siamese network~\citep{chopra2005learning}, which uses discriminatively learned features and refines them using a loss attached to the encodings of multiply winged networks over the compared images.
A similar version was later also developed which just uses the oracle triplets as supervision instead of refining a supervised version of the features, which is a setting we also consider~\citep{hadsell2006dimensionality}. Similar approaches have been used in~\cite{wang2014learning,schroff2015facenet}, where usage of supervised features with crowd-inferred similarities boosts performance in face verification and more generic fine-grained visual categorization tasks.
The key difference to our work is two-fold: we focus on a probabilistic encoder-decoder approach, where features are learned from images without labels and image information is not thrown away. Feature learning is guided additionally by an oracle and we introduce a probabilistic generative model which provides a joint model of all these components and their interactions, including semantic masking. This forces our model to learn explicit latent factors which capture the knowledge from the oracle rather than learning to be invariant to it and thus constitutes a harder and more comprehensive task.

Bayesian generative models have been proposed before for crowd-sourcing tasks~\citep{liu2012variational}. Our model differs in that we introduce latent variables generating the observations on which we evaluate triplet constraints, which are the observation from the oracle. Our setup better facilitates implicit knowledge transfer via posterior regularization.
Generative models in representation learning have recently made rapid progress using variational inference~\citep{kingma2013auto,rezende2014stochastic,mnih2014neural}. These techniques allow fast learning of directed graphical models and have been a major stepping stone in combining deep learning with graphical models. We briefly review variational autoencoders in Section~\ref{section:vae}. Notably, in ~\cite{kingma2014semi} these approaches are used to achieve state-of-the art results in semi-supervised learning with explicit labels. We identify that as a related setting to ours: using an oracle we can obtain weak implicit supervision in the form of similarity constraints over a sparse subset of the data and generalize from that, while most of the data is not subject to oracle constraints. Furthermore, we similarly can learn functionally compact subspaces which have semantic roles during generation.
In~\cite{cheung2014discovering}, deep generative models are used with functional constraints on the latent space to increase specificity of latent variables, which is a goal we share but tackle using the oracle-information as a model-based semantic regularizer.
Disentangling information and structuring models semantically is a theme in more recent work~\citep{reed2014learning},~\citep{kohli_graphicsNetworks}. 
Constraints on latent variables models in an otherwise unsupervised setting have also found early usage in the context of Gaussian Processes~\citep{lawrence2006local} using backconstraints.

Using side-knowledge as a regularizer for the posterior over latent variables has been explored in other settings for simpler latent variable models in~\cite{ganchev2010posterior} and we take inspiration from that work.
An interesting link also exists between our formulation of the triplet likelihood using the Jensen Shannon divergence and generative adversarial networks \citep{goodfellow2014generative}. The Bernoulli likelihood we employ using a softmax can conceptually be adapted to use a classifier to match the framework from~\cite{goodfellow2014generative}.

Finally, an intuitive connection also exists with Vapnik's privileged learning framework~\citep{vapnik2009new} where in a supervised setting improved classifiers can be learned if privileged information in the form of additional features is present during training time. Borrowing terminology, we consider the similarity constraints to be a sparse privilege conveyed by an oracle of unobserved structure and aim at learning a student model which improves understanding of the data. Our generative interpretation of this setting ultimately leads to our approach of learning a pseudo-causal inverse model of the data guided by oracle information by modeling factors of variation, instead of learning invariances as in~\cite{hadsell2006dimensionality,chopra2005learning}.

\vspace*{-1ex}
\section{Results}
\label{sec:results}
The aim of this section is to illustrate the key properties of our proposed algorithms. We start by describing the dataset and preliminaries in Section~\ref{sec-results-prelim}. In the first part (Section~\ref{sec-results-quant}), we quantitatively compare our methods against other baseline and state-of-the-art methods. In the second part (Section~\ref{sec-results-factorization}), we illustrate how our model variant with masks factorizes the latent spaces into distinct semantic units.
\vspace*{-1ex}
\subsection{Preliminaries}
\label{sec-results-prelim}
\vspace*{-1ex}
We use a relatively small dataset that is, however, well-suited to illustrate the features of the algorithm and facilitates the interpretation of the factorized latent spaces:  the Yale Faces dataset~\citep{lee2005acquiring}. The version we used comprises of $2,414$ images from $38$ individuals under different light conditions. We split it into $300$ test images and $2,114$ training images. The images were taken under controlled conditions using a lighting rig which allows for light sources to be varied in specific ways. The azimuth and elevation of the light in relation with the depicted face were changed with values between $-130$ to $+130$ degrees and $-40$ to $90$ degrees, respectively. The resulting images have dramatic variability in appearance due to shading, apart from variability in identity of the depicted person.


We proceeded with a series of  oracle simulations. Particularity, we simulate three different questions upon presenting it with random triplets of images which we will then use the evaluations below. The questions we used were  the following: 
\begin{enumerate}
\item Who has the most similar identity? ("Identity")
\item Where is the light condition most similar in terms of azimuth? ("Azimuth")
\item Where is the light condition most similar in terms of elevation? ("Elevation").
\end{enumerate}
While the first question is similar to a typical classification setting, answering it accurately may actually require the ability to understand light variation well. Question 2 \& 3 concern  complex qualities of the images related to visual physics. The used Yale faces dataset provides metadata for all images and we can use this metadata to simulate a number of triplets for each oracle.

\vspace*{-1ex}
\subsection{Comparison with baseline and state-of-the-art methods}
\label{sec-results-quant}
\vspace*{-1ex}

For evaluation we aim at a more quantitative understanding of what our and other models are good at and where they have limitations. For this we consider the metric learning network  analogous to~\cite{hadsell2006dimensionality} and the a purely unsupervised variational autoencoder (see Section~\ref{section:vae}) as comparators. We refer to them as {\bf MetricL} and {\bf VAE}, respectively. VAE is independent of oracle triplets in all experiments since it works entirely unsupervised. 


We use the learned representations of each model to assess the quality with respect to different evaluation measures. In particular, we use the representations to predict the identity of the face, the azimuth degree and the elevation degree (we provide classification error and RMSD for the degrees). The latter is to assess how well the models capture physical properties of the images.  All evaluations are done on held-out test data using a logistic regression model. In addition, we measure how well each model is able to predict triplets on test data, i.e., predict whether triplet $i,j,l$ or $i,l,j$ is true. We provide further information for model details and experimental setup in Appendix~\ref{sec:appA}.

%


\begin{table}[b]
\vspace*{-3ex}
\begin{subtable}{.495\textwidth}
\scalebox{0.95}{\begin{tabular}{|l|c|c|c|}
\hline
Oracle/Method         &	MetricL & VAE & OPBN \\
\hline
\underline{Identity}  &	{\bf 9.0}	 & 18.2	      & {\bf 9.0}  \\[0.2ex]
Azimuth	              & 36.3	     & {\bf 20.4} & {\bf 20.6} \\[0.2ex]
Elevation	          & 20.0	     & {\bf 10.5} & {\bf 10.4} \\[0.2ex]
\hline		
Triplet prediction	  & {\bf 1.25}	& 34.00	 & \emph{6.42} \\
\hline
\end{tabular}}
\caption{Model trained with the identity oracle. \label{table_id_oracle}}\vspace*{-1ex}
\end{subtable}
\begin{subtable}{.495\textwidth}
\scalebox{0.95}{\begin{tabular}{|l|c|c|c|}
\hline
Oracle/Method         &	MetricL  & VAE        & OPBN \\
\hline
Identity              &	70.0	   & {\bf 18.2} & {\bf 18.7}  \\[0.2ex]
\underline{Azimuth}	  & {\bf 13.5} & 20.4       & \emph{16.4} \\[0.2ex]
Elevation	          & 18.0	   & {\bf 10.5} & {\bf 10.2} \\[0.2ex]
\hline		
Triplet prediction	  & {\bf 3.4}& 34.00	  & \emph{6.60} \\
\hline
\end{tabular}}
\caption{Model trained with the azimuth oracle. \label{table_az_oracle}}\vspace*{-1ex}
\end{subtable}
\caption{Comparison of Metric Learning Networks (MetricL), Variational Autoencoders (VAE) and our proposed model without masks (OPBN). We train the model with 100,000 triplets from the identity oracle (left) and the azimuth oracle (right). Best results results are in bold face. Second best results italic. We observe that OPBN predicts all properties reasonably well, while MetricL only for the task it is trained for. VAE works well on predicting lighting conditions. For more details see main text.\label{table-no-masks}\vspace*{-3ex}}
\end{table}

The results reveal that {\bf OPBN} and its variants are on average the best-performing method. MetricL effectively learns a classifier in this setting. In Table~\ref{table_id_oracle} we see that the generative model competes with the metric learning method in terms of classification when being informed about identity from the oracle, while maintaining low error rates for tasks related to image physics. It outperforms VAE on classification accuracy, at no loss to image understanding. We believe this shows that it incorporates oracle knowledge to shape alternative latent spaces compared to VAE.
In Table~\ref{table_az_oracle} we test how well a model can incorporate more subtle oracle knowledge. We inform models only using a light condition oracle (azimuth). As expected, the metric learning performance collapses on all tasks except on the targeted oracle-task. VAE maintains the same performance since it is agnostic to oracle information. OPBN on the other hand maintains good performance on all tasks, benefits in predicting light conditions over the unsupervised VAE.

In a more complex setting, we also give the models oracle-information from all available questions jointly and test their performance. As we show in Table~\ref{table-masked-opbn}, again OPBN is the best-performing method on average. Metric learning approach cannot incorporate the variability of the available information usefully when using just a few triplets (data not shown), but in the setting we report using 100000 triplets it achieves good performance on classification and elevation prediction. The main difference in performance between VAE and OPBN is in classification and the ability to predict triplets, which correlates with our observation that only about 50\% of training triplets would be satisfied with the VAE approach. We see, that OPBN learns an equally competitive, but clearly different latent space than VAE and captures the semantics of the oracle better by being more predictive on unseen triplets on test data.
However, the benefits of the oracles when incorporating multiple queries are underwhelming in comparison to single oracles. To address this issue, we add an experiment using Masked-OPBN. We observe that masked OPBN exhibits greatly improved quantitative performance on all tasks and yields representations which are more predictive of the image-properties, the class and the held out triplets than the other models. 
To further test this capacity after seeing that OPBN stalls in its improved performance, we add an extra experiment with 5 times the triplets for Masked OPBN. We observe that masking allows the model to continue improving as more triplets are added. OPBN with masks thus shows the greatest promise to incorporate heterogeneous information from oracles into its latent spaces.


\begin{table}[t]
\begin{center}
\vspace*{-4ex}\begin{tabular}{|l|c|c|c|c|c|}
\hline
Oracle/Method         & MetricL	& VAE      & OPBN           & OPBN Mask   & OPBN Mask 500k \\
\hline
\underline{Identity}  & 10.2	&  18.2    &  \emph{9.7}    & \emph{9.0}  & {\bf 7.3} \\
\underline{Azimuth}	  & 27.1	& 20.4	   & 19.8           & {\bf 13.8}  & {\bf 14.1} \\
\underline{Elevation} & 9.1	    & \emph{10.5} & \emph{10.0} & {\bf 6.5}   & {\bf 6.0} \\
\hline					
Triplet prediction	  & 27.1$^\ast$	& 34.0$^\ast$ & \emph{5.7} & \emph{5.2}  & {\bf 4.2} \\
\hline
\end{tabular}\vspace*{-2ex}
\end{center}

\caption{Comparison of Metric learning networks (MetricL), Variational Autoencoders  (VAE) and three variants of our model (OPBN in Figure~\ref{table-no-masks}, OPBN with masks and OPBN with masks and 500,000 triplets instead of 100,000 triplets per oracle). We use triplets from the three oracles Identity, Azimuth and Elevation. We observe that OBPN with masks performs much better on all evaluations. Using more triplets leads to further improvements. Numbers marked with $\ast$ should be considered with care since MetricL and VAE are not aware of differences between oracles. See main text for more details.  \label{table-masked-opbn}\vspace*{-4ex}}
\end{table}

\vspace*{-1ex}
\subsection{Masking OPBN Leads To Factorization of Latent Spaces}
\label{sec-results-factorization}
\vspace*{-1ex}

\begin{figure}[t]
\vspace*{-4ex}\includegraphics[width=\textwidth]{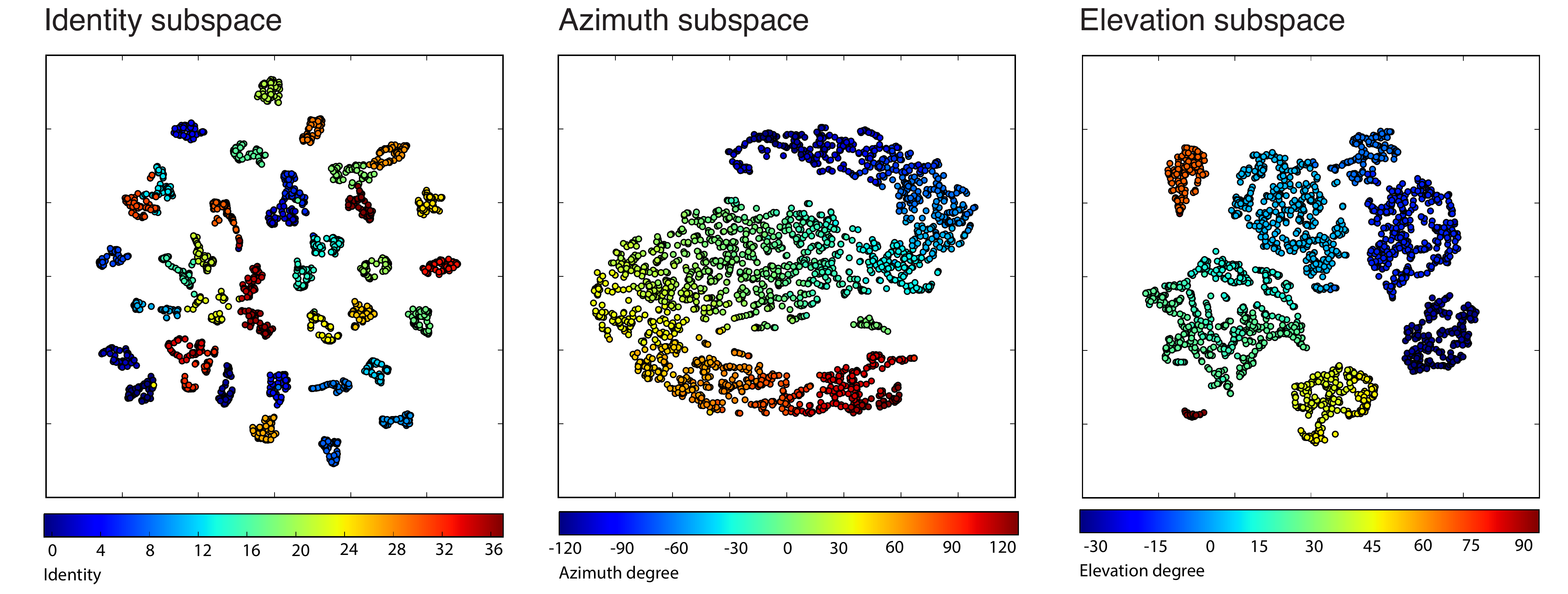}
\caption{t-SNE visualizations of the latent subspaces that were identified in the model with shown weights in Figure~\ref{fig_masks}. For visualization, we only use the dimensions with weight greater than $0.2$ for each oracle. We observe that the identity subspace clearly separates faces from different persons (left), the azimuth degree (middle) and elevation degree (right) of light exposure.\label{fig-embedding} \vspace*{-2ex}}
\end{figure}

\begin{figure}[t]
\parbox{0.4\textwidth}{\includegraphics[width=0.36\textwidth]{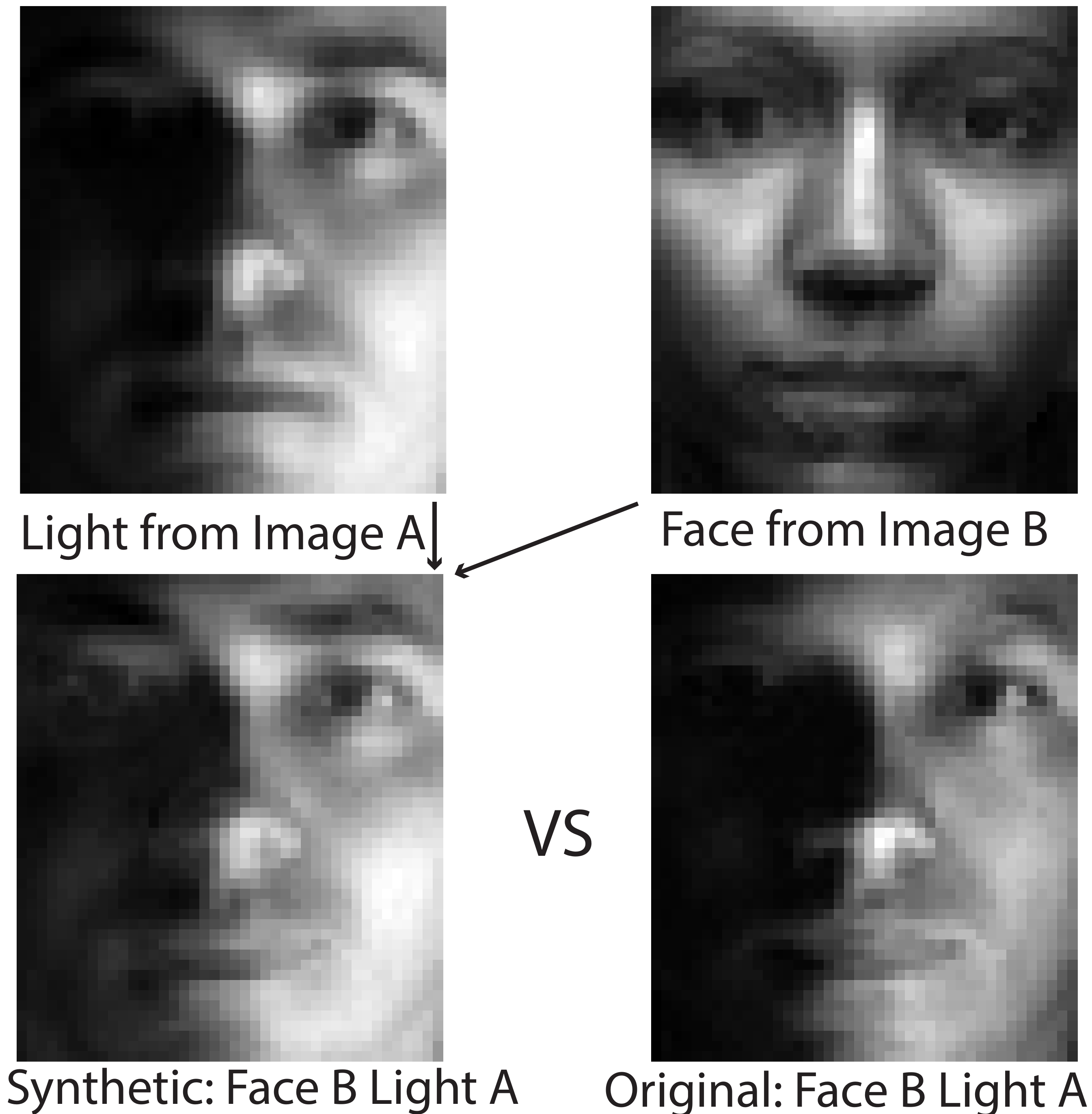}}%
\parbox{0.6\textwidth}{\caption{We illustrate the factorization of latent spaces when using masked models in the following. We take two images from the training-set, A and B, and project them into latent space. We use the encoding for face given by the identity mask from image B and combine it with the latent features given by the azimuth mask applied to the encoding of image A. The resulting image is a blend of  two, as expected, and approximates the facial features of image B, especially the mouth region and facial shape. The blended image furthermore exhibits light properties similar to image A. We finally show an unobserved test image which shows face B under light conditions A for comparison. We see that the facial transfer is not perfect, as the eyebrows are still taken from image A and the skin shading is a blend of both images. These are fair mistakes, since eyebrows are frequently shaded or mixed up with to light conditions in this dataset. We expect this to improve on bigger datasets and when using more oracle samples.}\label{fig-faces}\vspace*{-5ex}}\vspace*{-2ex}
\end{figure}

\begin{figwindow}[0,r,{\includegraphics[width = 2.6in]{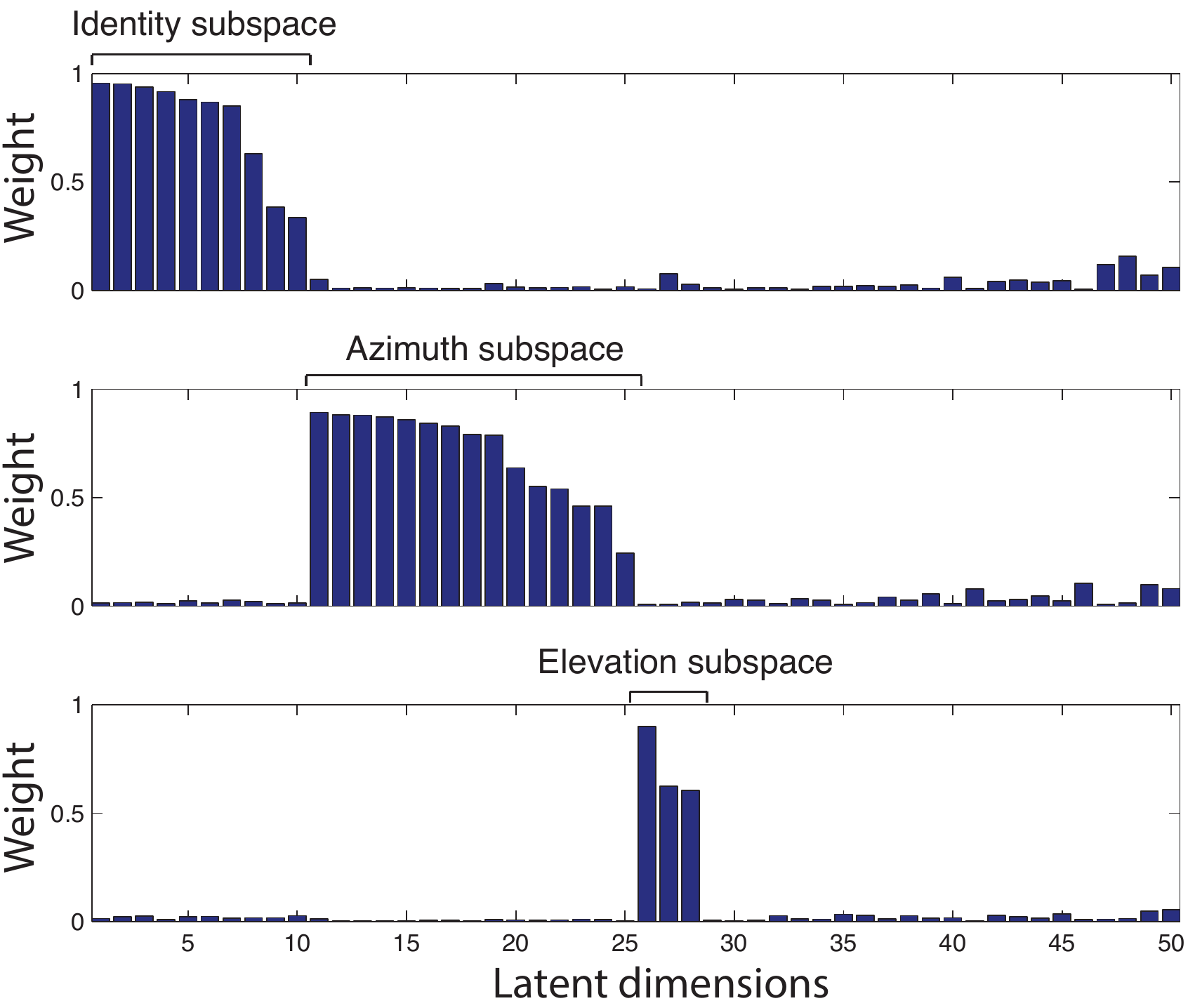}
},{We show the oracle-specific masks over the latent space when learning from multiple oracles at once. It is evident, that the model learns to virtually switch off different dimensions per question and learns to factorize the latent space into compartmentalized, task-relevant subspaces without ever explicitly receiving supervision, for instance, about how to factorize light from a face class.\label{fig_masks}}]
We observed that the masked version of OPBN shows greatly improved ability to learn from complex oracles with multiple heterogeneous queries compared to  other discussed approaches (see Figure~\ref{table-masked-opbn}).
For a model with otherwise equal parametric capacity and the same fundamental inference machinery, this constitutes a surprising observation. In this section, we will illustrate the effects of the learned masks and how they contribute to performance improvements.

In the non-masked models triplet likelihoods are global. By learning local likelihoods via masking, or subspaces, we allow the model to decide which parts of the space it uses per query. Variational compression leads to solutions with the least possible amounts of used variables. In Figure~\ref{fig_masks} we show a fully learned mask for the Yale Faces model for each query. Learning jointly from all queries leads to a factorization of the latent space into task specific and partially shared latent variables. We also observed a strong quantitative footprint from usage of these masks: performance on all predictive tasks for models of otherwise equal capacity improves across the board (see Table~\ref{table-masked-opbn}), leading to models which capture light conditions and class better jointly. We consider this an effect of knowledge transfer from the oracle/crowd, allowing the model to identify semantic latent variable systems which do not only strive for high likelihoods on pixels, but also help model oracle triplets. We also observe that models with masks improve dramatically with availability of more triplets. The Bayesian objective helps compress the latent spaces into semantic variables. 

In order to inspect the latent spaces induced by the masks, we embed the respective subspaces of the latent variable encoding using t-SNE in Figure~\ref{fig-embedding}. It reveals that we learn fine-grained class clusters when using the identity subspace, and continuous and smooth embeddings for azimuth and elevation pointing to the understanding of the model of the continuous and nature of light placements in the images. These results point to the fact that the oracle-informed model is able to learn the semantics of light placement and of facial structure in dedicated subspaces, increasing the semanticness of the learned space significantly. This helps identify semantic variables which were explicitly never observed, as sketched earlier in Figure~\ref{fig_systemid}.
We finally present an example of using masks to sample synthesized images in Figure~\ref{fig-faces}. This illustrates the controlled transfer of subtle imaging-physics properties from one image to the next using the model.

\end{figwindow}

\vspace*{-1ex}
\section{Discussion}
\vspace*{-1ex}
\label{sec:discussion}
We have introduced a joint unsupervised generative model over observations and triplet-constraints as given by an oracle. Our contributions are first a fully probabilistic treatment of triplets and latent variable models in a joint unsupervised setting using variational belief networks. We show how this joint learning allows for implicit knowledge from an oracle, such as a human crowd, to be transferred to a rich parametric model, resulting in improved classification scores, improved ability to predict triplets and more interpretability of the crowd biases. This can be a useful framework to encode expert knowledge in probabilistic reasoning systems when the exact model is unknown or labels are hard to obtain.
Second, we introduce information theoretic distance measures for triplets generalizing the commonly used Euclidian distances.
We furthermore introduce the notion of question specific masks in latent space to force the model to identify interpretable features of relevance for each specific type of oracle constraint, enabling the model to learn from multiple types of questions at once and boosting performance further.
Our approach using variational inference and a triplet likelihood is not limited to belief networks, thus it will be interesting to use the framework in conjunction with other flexible probabilistic models such as Gaussian Processes and infinite partition models.
We highlight the fact that using our framework no supervised pre-training of features is needed, as it can learn problem specific nonlinear feature-spaces adapted to the available information.
We showed that our approach compares favorably with state-of-the-art metric learning models and fully unsupervised method in a generic application using feedforward networks. Our model is trivially extendable with convolutional and de-convolutional networks to be used on high-dimensional data.
It will be interesting to combine the learning approach with more structure in temporal or spatially constrained models and encode other relationships like topological or unobserved constraints, such as taste of food in images.
On the oracle side, future work regarding more accurate crowd-modeling for different bias and noise regimes are promising in conjunction with use-cases such as amazon mechanical turk.
Our model is also particularly amenable to active learning for probing the oracles optimally.
Finally, we wish to mention the potential for this framework to assist perceptual applications where biases of the human visual system can be studied assisted by generative models.

\newpage
\bibliography{crowd_vae}
\bibliographystyle{iclr2016_conference}

\begin{appendix}
\newpage
\section{Appendix A: Experimental Settings}
\label{sec:appA}
Here we can give more experimental details.

In all experiments we used diagonal Normal distributions as priors for the latent space and {\sl rmsProp} with momentum~\citep{graves2013generating} or {\sl ADAM}~\citep{kingma2014adam} as an optimizer.
All experiments were run on Graphics Processing Units (GPUs) using a {\sl theano}~\citep{bergstra+al:2010-scipy} implementation and did not take more than a few hours each.

We can simulate an oracle for each question by using the annotations provided with the dataset. For question one, we sample from the label distribution checking for a match to produce answers to the triplets generated. For question two and three we resort to sampling from the relative distances of target angles to given angles to produce the triplet information.
We finally generate 3 different simulated oracles, OracleID, OracleAz and OracleAll which correspond to asking just the first question, just the second or all three mixed.
We sample 100,000 triplets for each oracle-question at random (meaning that the third simulation has 300000 triplets). We repeat the process 3 times to account for sampling bias and report the means of the reruns on our experiments. In an extra illustrative example combining all three oracles, we samples 500,000 triplets per question for a total of 1,500,000 triplets to inform the model. While these numbers sound high, we note that there is a large combinatorial space of possible triplets to be explored.

We proceed to learn fully unsupervised models of these images using an architecture with 200 hidden deterministic units and 50 latent variables. The deterministic layers use \emph{tanh} nonlinearities. We set up the analogous {\bf MetricL} model without a generative path as a supervised learning model optimizing the triplet embeddings given images with a euclidian loss function.

\section{Appendix B: Details for Oracle-Likelihood}
\label{sec:appB}
In order to compute the likelihood for the triplet likelihood, we need to calculate an expensive divergence term $D$ using an information theoretic quantity, the Jensen Shannon Divergence as defined in Section~\ref{sec:triplet_model}. In practice, this term is typically intractable analytically since it involves a KL divergence involving over a mixture over two possibly disjoint distributions. In order to evaluate this KL divergence, exhaustive sampling methods need to be used.

In order to avoid expensive sampling steps during training, we explore approximations to the term $D$. In the presented experiments we used:
\begin{equation*}
D_{a,b} = \sum \limits_{h=1}^{H} D^{h}_{a,b}=-\sum \limits_{h=1}^{H} \left[ \frac{1}{2} \text{KL}\Big(p(\z_{a}^{h})||p(\z_{b}^{h})\Big) +\frac{1}{2} \text{KL}\Big(p(\z_{b}^{h})||p(\z_{a}^{h})\Big)\right].
\label{eq_D12a}
\end{equation*}
This approximation is inaccurate globally, but empirically is fast and yields better results than the KL divergence or a eucilidian distance and becomes accurate in the limit of closeby distributions.
Clearly, using the full JS is beneficial to the model and yields stronger posterior regularization allowing to learn more efficiently from triplets, especially in combination with full covariance latent spaces.
An overview of previous approximations related to the JS is given in \citep{hershey2007approximating}.
We have tried previously known Monte Carlo-based approximations and explore novel deterministic approximations to this term and expect to show empirical performance in an update to this paper and in follow-up work.

\section{Appendix C: Further Yale Faces Samples}
\label{sec:appC}

We trained a model on Yale Faces with 400 hidden units (units chosen until likelihoods stopped improving) and used it similarly to the masked experiments in the main paper. We use the space in the supplement to show a few more samples in Figure~\ref{fig:yale_samples} and do a form of image algebra by adding components of various images together.

\begin{figure}[t]
\includegraphics[width=\textwidth]{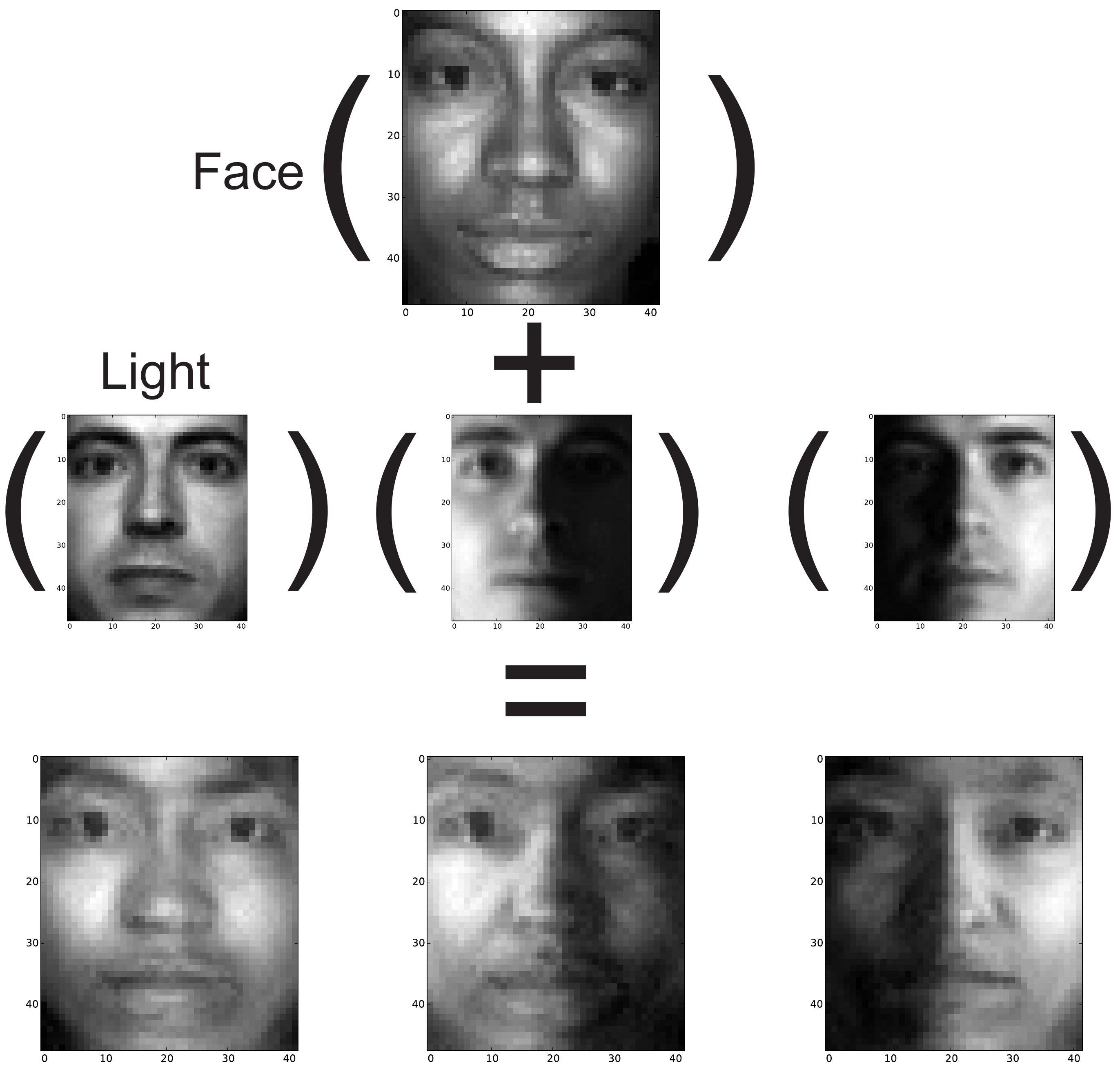}
\caption{ {\bf Top} We select a trainings image of a face and select the latent encoding corresponding to its identify. {\bf Middle} We select three training images of another face under different light conditions and select the light variables according to the mask. {\bf Bottom} We synthesize new images with the face and light images clamped to the observations and see that noisy faces are generated which look like the top face and have light conditions like the middle one. }
\label{fig:yale_samples}
\end{figure}

\newpage
\section{Appendix D: Yale Faces Triplet Variations}
\label{sec:appD}

We sample another batch of 100000 triplets for the Yale dataset per query and rerun OPBN-Masked with a varying number of triplets to clarify the effect. We show results in Table~\ref{table_yale_triplet_vary}, where it is evident that all queries improve as we add triplets. 
We want to note that these numbers are based on a single sampling of triplets and thus are subject to sampling noise. By chance, more or less good triplets may be contained in the set.

\begin{table}[b]
\caption{Comparison of Model metrics on Yale between with varying triplet numbers.}
\label{table_yale_triplet_vary}
\begin{center}
\begin{small}
\begin{sc}
\begin{tabular}{lcccr}
\hline
OracleAll & 100 & 1000 & 10000 & 100000  \\
\hline
Triplet Prediction& 35.95 & 28.88 & 22.95 & {\bf 5.56} \\
Classification    & 19.00 & 15.66 & 12.66 &  {\bf 8.66} \\
Azimuth RMSD      & 19.59 & 18.73 & 17.02 &  {\bf 15.50} \\ 
Elevation RMSD    & 9.59  & 10.99 & 7.75 &  {\bf 6.37} \\ 
\hline
\end{tabular}
\end{sc}
\end{small}
\end{center}
\end{table}

\section{Appendix E: MNIST Experiments}
\label{sec:appE}
\begin{figure}[t]
\includegraphics[width=\textwidth]{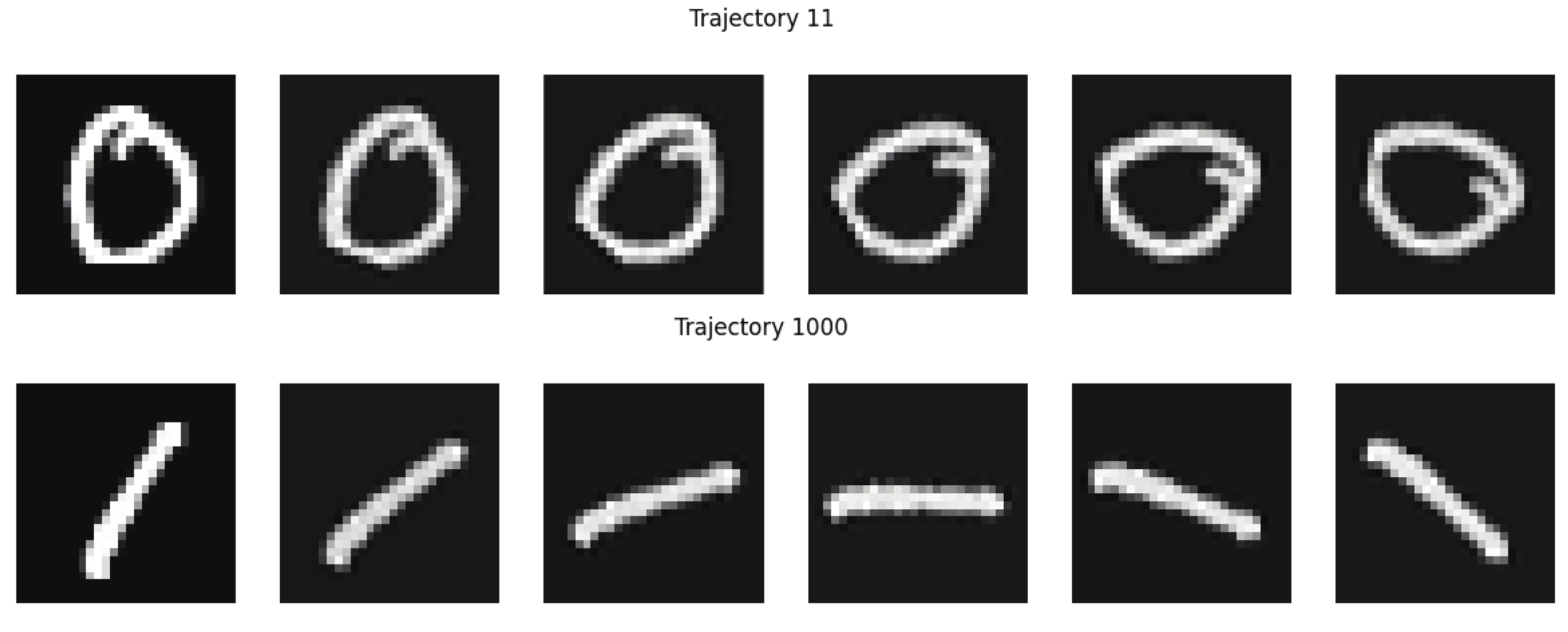}
\caption{We show the generated training data. {\bf Left} the original image {\bf Right} progressively rotated version of the original, depending on position on trajectory.}
\label{fig:mnist}
\end{figure}

We generated a perturbed version of the MNIST dataset in order to show other settings where the proposed approach can be used interestingly.
In normal MNIST, each letter is generated depending on its class. By eyeballing, style variations can be seen, but they are not captured in the meta-data in order to be used for evaluation. We proceed to take 10000 MNIST digits of equal proportions from each class and rotate them by 5 progressively increasing positive angles. This creates the effect of pushing the digits to fall over towards the right side in a trajectory, as shown in Figure~\ref{fig:mnist}. The questions we can ask simulated oracles here are the following:
\begin{enumerate}
\item Which image of a set is part of the same trajectory? This question is related to both identity and instantiation (style) of a variable.
\item Which images have similar/dissimilar angles/timepoints?
\item Which images have similar labels?
\end{enumerate}

We chose not to include order into the labels, but similarity in digit-images could also be defined by the value of a digit which could be used for reasoning tasks such as performing mathematical operations with inferred values in an ordered manifold. In our setting we assume similarity is implied by the same label and dissimilarity else. In future work we plan to exploit semantic oracles which understand order for reasoning-related tasks.
In our experiment using 2 deterministic layers with 800 and 400 hidden units with tanh nonlinearities and 50 latent variables we find OPBN masked to outperform VAE, see Table~\ref{table_mnist_all}. The mask variables also manage to factorize the latent space sharply into variables for each query in the setting of 100000 triplets. We also observed that when using less triplets the second query suffered the most in performance, whoch makes intuitive sense since learning a rotation is a harder task than learning to match labels. In terms of final performance we observe that OPBN can strongly reduce the predictive errors for the tasks, although the two synthetic tasks are actually quite hard.

\begin{table}[h]
\caption{Comparison of Model metrics on MNIST between VAE and OPBN masked with 100, 1000 and 100000 triplets.}
\label{table_mnist_all}
\vskip 0.15in
\begin{center}
\begin{small}
\begin{sc}
\begin{tabular}{lcccr}
\hline
OracleAll & VAE  & OPBNmasked-100 & OPBNmasked-1000 & OPBNmasked-100000  \\
\hline
Triplet Prediction  & 36.51 & 34.35 & 34.32 & {\bf 16.45} \\
Classification  & 24.19 & 23.05 & 22.61  &  {\bf 10.68} \\
Rotation Angle RMSD & 18.87 & 18.4 & 18.7 &  {\bf 14.28} \\ 
\hline
\end{tabular}
\end{sc}
\end{small}
\end{center}
\vskip -0.1in
\end{table}

\newpage
\section{Appendix F: Robustness to Oracle-Noise}
Here we train Metric Learning and OPBN with 2,000 triplets and 2,000 datapoints in a variety of oracle settings.
Oracles are perturbed by noise $\epsilon$, meaning that a fraction equal to $\epsilon$ of the triplets are flipped and thus wrong. The experiment illustrates the robustness the generative aspect of ghe model gives it, whereas it is evident that metric learning approaches lose more performance since they cannot benefit from modeling observations directly.  We observe that OPBN learns significantly better representations to predict azimuth, elevation and the classification label. OPBN performs similarly good in terms of triplet prediction.

\begin{table}
\caption{Comparison of Model metrics on Yale Faces with Identity crowd.}
\label{complexity_table:id}
\vskip 0.15in
\begin{center}
\begin{small}
\begin{sc}
\begin{tabular}{lcccr}
\hline
OracleID $\epsilon=0$  & MetricL & OPBN \\
\hline
Azimuth RMSD&  57 & 20.99 \\
Elevation RMSD& 24.3 & 10.13\\
Classification\% & 14 & {\bf 7.2}\\
Triplet Prediction \%  & {\bf 5.0} & 5.3\\
\hline
\end{tabular}
\end{sc}
\end{small}
\end{center}
\vskip -0.1in
\end{table}

\begin{table}
\caption{Comparison of Model metrics on Yale Faces with Azimuth crowd}
\label{complexity_table:az}
\vskip 0.15in
\begin{center}
\begin{small}
\begin{sc}
\begin{tabular}{lcccr}
\hline
OracleAZ $\epsilon=0$ & MetricL & OPBN \\
\hline
Triplet Prediction \% & {\bf 10.8} & 18.1\\
Azimuth RMSD&  {\bf 18} & 19.8 \\
Elevation RMSD &18.59 & 11.7\\
Classification \%& 40.7 & {\bf 7.9}\\
\hline
\end{tabular}
\end{sc}
\end{small}
\end{center}
\vskip -0.1in
\end{table}

\begin{table}
\caption{Comparison of Model metrics on Yale Faces with All-oracle.}
\label{complexity_table:all}
\vskip 0.15in
\begin{center}
\begin{small}
\begin{sc}
\begin{tabular}{lcccr}
\hline
OracleAll $\epsilon=0$ & VAE & MetricL & OPBN & OBPN-Masked \\
\hline
Triplet Prediction \% & 38.7 & 33 &  18.4 & {\bf 9.7}\\
Azimuth RMSD&  21.18 & 40.42 & 21.77 &  {\bf 16.37}\\
Elevation RMSD& 11.57 &14.75 &  10.23 & {\bf 7.14}\\
Classification \% & 17 & 25 & {\bf 12.6} &  13.6 \\
\hline
\end{tabular}
\end{sc}
\end{small}
\end{center}
\vskip -0.1in
\end{table}

\begin{table}
\caption{Comparison of Model metrics on Yale Faces for noise noise robustness when using MetricL.}
\label{complexity_table:metricNoise}
\vskip 0.15in
\begin{center}
\begin{small}
\begin{sc}
\begin{tabular}{lcccr}
\hline
MetricL-All & $\epsilon=0$ &  $\epsilon=0.2$ & $\epsilon=0.4$ \\
\hline
Triplet Prediction \% & 67 & 59.9 & 52\\
Azimuth RMSD& 40.42 & 35.32 & 41.42 \\
Elevation RMSD& 14.75 &18.62 & 21.6\\
Classification \% & 75 & 71 & 71\\
\hline
\end{tabular}
\end{sc}
\end{small}
\end{center}
\vskip -0.1in
\end{table}

\begin{table}
\caption{Comparison of Model metrics on Yale Faces for noise robustness when using OPBN.}
\label{complexity_table:OPBNnoise}\label{complexity_table:metricNoise}
\vskip 0.15in
\begin{center}
\begin{small}
\begin{sc}
\scalebox{0.8}{\begin{tabular}{lcccr}
\hline
OPBN & $\epsilon=0$ &  $\epsilon=0.2$ & $\epsilon=0.4$ \\
\hline
Triplet Prediction \% & 18.4 & 37.6 & 46.2\\
Azimuth RMSD& 21.77 & 22.9 & 23.75 \\
Elevation RMSD& 10.23 &10.89 & 10.72\\
Classification \% & 92.6 & 89.3 & 89.1\\
\hline
\end{tabular}}%
\scalebox{0.8}{\begin{tabular}{lcccr}
\hline
MetricL-All & $\epsilon=0$ &  $\epsilon=0.2$ & $\epsilon=0.4$ \\
\hline
Triplet Prediction \% & 67 & 59.9 & 52\\
Azimuth RMSD& 40.42 & 35.32 & 41.42 \\
Elevation RMSD& 14.75 &18.62 & 21.6\\
Classification \% & 75 & 71 & 71\\
\hline
\end{tabular}}
\end{sc}
\end{small}
\end{center}
\vskip -0.1in
\end{table}

\end{appendix}
\end{document}